\documentclass[runningheads]{llncs}
\usepackage{graphicx}
\usepackage{comment}
\usepackage{amsmath,amssymb} % define this before the line numbering.
\usepackage{color}

\usepackage{mathrsfs}
\usepackage{multirow}
\usepackage{caption}
\usepackage{graphicx}
\usepackage{amsmath}
\usepackage{bm}
\usepackage{marvosym}

\newcommand{\etal}{\textit{et al}.}
\newcommand{\ie}{\textit{i}.\textit{e}.}
\newcommand{\eg}{\textit{e}.\textit{g}.}

\usepackage{hyperref}
\makeatletter
\newcommand{\printfnsymbol}[1]{%
	\textsuperscript{\@fnsymbol{#1}}%
}
\makeatother

\begin{document}
	\title{RGB-D Salient Object Detection with Cross-Modality Modulation and Selection} % Replace with your title
	% Replace with your title
	
	\titlerunning{RGB-D SOD with Cross-Modality Modulation and Selection}
	% Replace with a meaningful short version of your title
	%
	\author{Chongyi Li\thanks{Equal contribution}\inst{1}\and
		Runmin Cong\printfnsymbol{1}\textsuperscript{\Letter}\inst{2} \and
		Yongri Piao\inst{3} \and
		Qianqian Xu\inst{4} \and \\
		Chen Change Loy\inst{1}}
	%
	%Please write out author names in full in the paper, i.e. full given and family names.
	%If any authors have names that can be parsed into FirstName LastName in multiple ways, please include the correct parsing, in a comment to the volume editors:
	%\index{Lastnames, Firstnames}
	%(Do not uncomment it, because you may introduce extra index items if you do that, we will use scripts for introducing index entries...)
	\authorrunning{Li \etal}
	% Replace with shorter version of the author list. If there are more authors than fits a line, please use A. Author et al.
	%
	
	\institute{School of Computer Science and Engineering, Nanyang Technological University, Singapore \and
		Institute of Information Science, Beijing Jiaotong University, Beijing, China \and
		School of ICE, Dalian University of Technology, Dalian, China  \and
		Institute of Computing Technology, Chinese Academy of Sciences, Beijing, China \\
		 %\email{lichongyi25@gmail.com, rmcong@bjtu.edu.cn}\\
		 %\email{yrpiao@dlut.edu.cn, xuqianqian@ict.ac.cn, ccloy@ntu.edu.sg}\\
		\url{https://li-chongyi.github.io/Proj_ECCV20}}
	\maketitle              % typeset the header of the contribution

\begin{abstract}
We present an effective method to progressively integrate and refine the cross-modality complementarities for RGB-D salient object detection (SOD).
The proposed network mainly solves two challenging issues: 1) how to effectively integrate the complementary information from RGB image and its corresponding depth map, and 2) how to adaptively select more saliency-related features.
\textit{First}, we propose a cross-modality feature modulation (cmFM) module  to enhance feature representations by taking the depth features as prior, which models the complementary relations of RGB-D data.
\textit{Second}, we propose an adaptive feature selection (AFS)  module to  select saliency-related features and suppress the inferior ones. The AFS module exploits multi-modality spatial feature fusion with the self-modality and cross-modality interdependencies of channel features are considered.
\textit{Third}, we employ a saliency-guided position-edge attention (sg-PEA) module to encourage our network to focus more on saliency-related regions.
The above modules as a whole, called cmMS block, facilitates the refinement of saliency features in a coarse-to-fine fashion.
Coupled with a bottom-up inference, the refined saliency features enable accurate and edge-preserving SOD.
Extensive experiments demonstrate that our network outperforms state-of-the-art saliency detectors on six popular RGB-D SOD benchmarks.
%\dots
%\keywords{RGB-D Images, Salient Object Detection, Multi-Modality Complementary, Feature Modulation and Selection}
\end{abstract}

\section{Introduction}

\begin{figure}
	\begin{center}
		\begin{tabular}{c@{ }c@{ }c@{ }c@{ }c@{ }c@{ }c}
			\includegraphics[width=.15\textwidth,height=1.5cm]{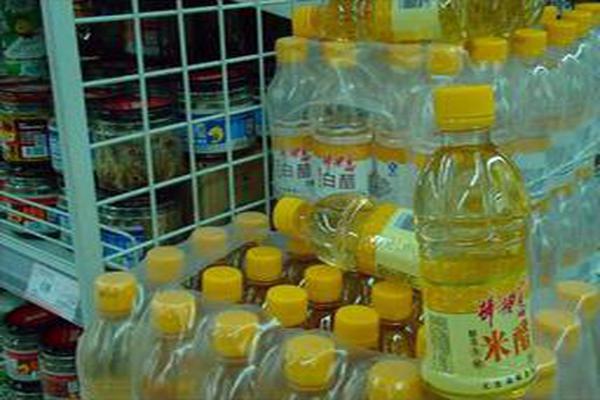}&
			\includegraphics[width=.15\textwidth,height=1.5cm]{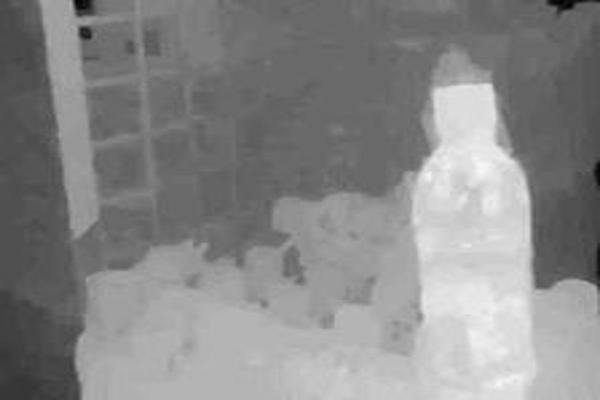}&
			\includegraphics[width=.15\textwidth,height=1.5cm]{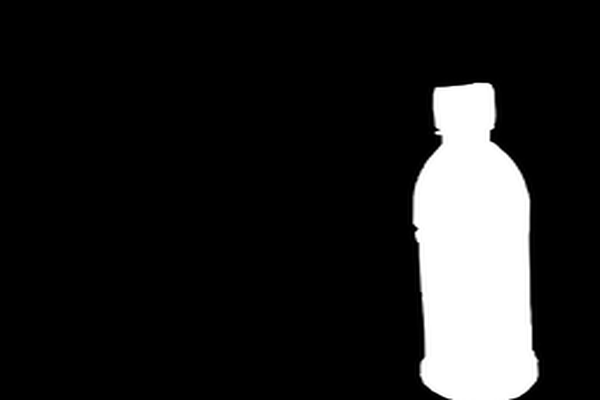}&
			\includegraphics[width=.15\textwidth,height=1.5cm]{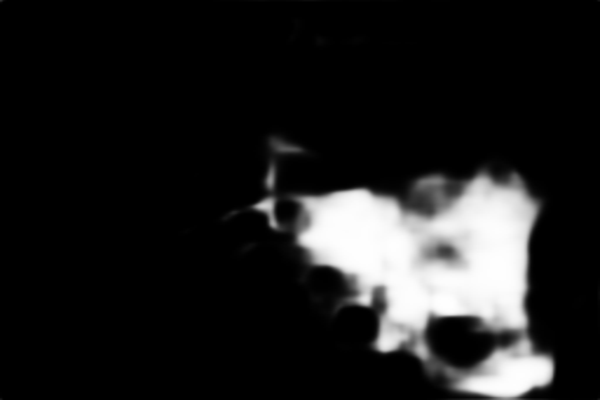}&
			\includegraphics[width=.15\textwidth,height=1.5cm]{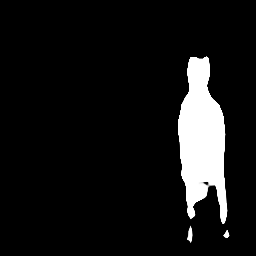}&
			\includegraphics[width=.15\textwidth,height=1.5cm]{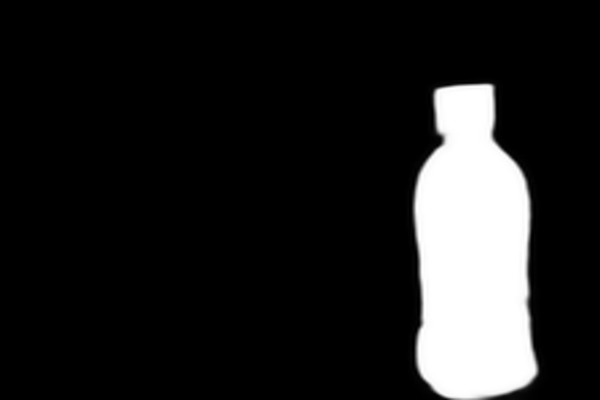}\\
			\includegraphics[width=.15\textwidth,height=1.5cm]{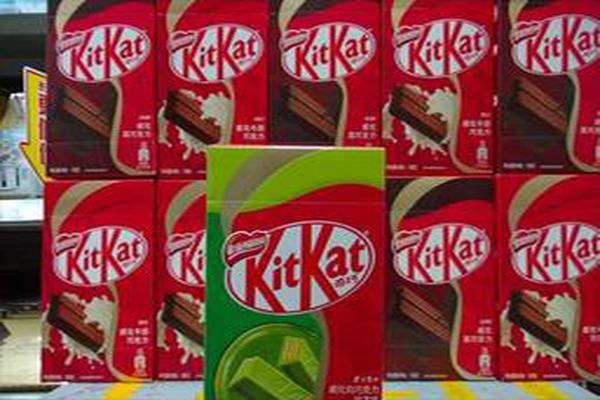}&
			\includegraphics[width=.15\textwidth,height=1.5cm]{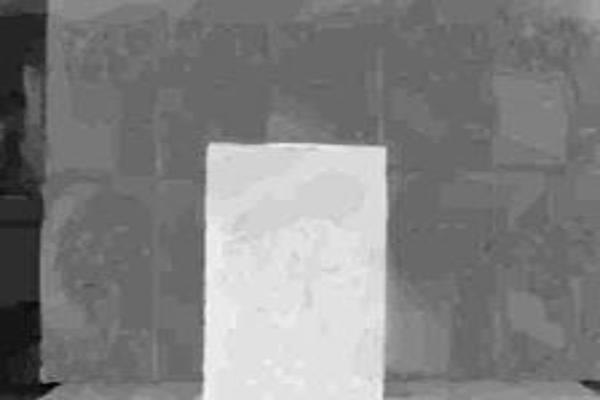}&
			\includegraphics[width=.15\textwidth,height=1.5cm]{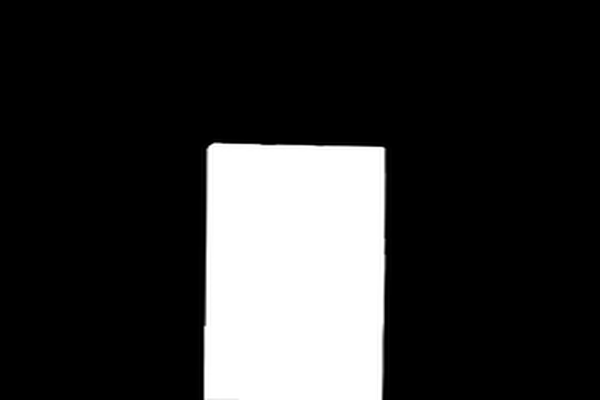}&
			\includegraphics[width=.15\textwidth,height=1.5cm]{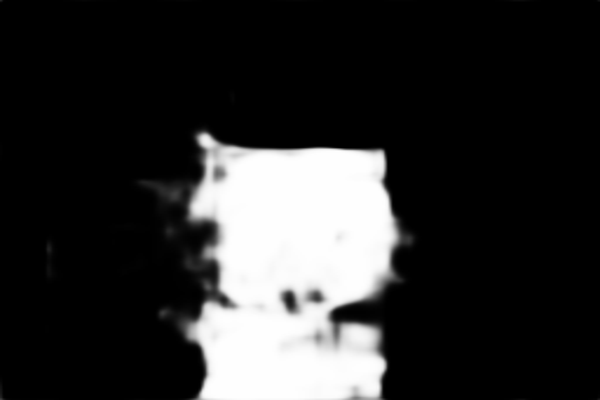}&
			\includegraphics[width=.15\textwidth,height=1.5cm]{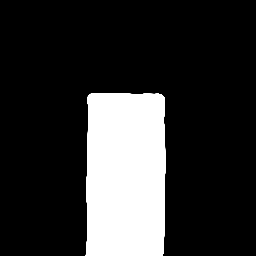}&
			\includegraphics[width=.15\textwidth,height=1.5cm]{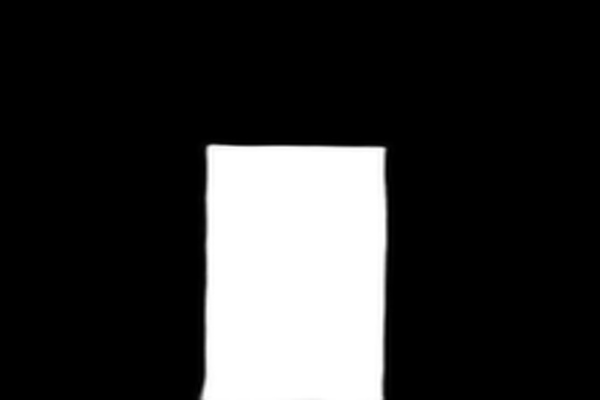}\\
			(a) RGB & (b) Depth & (c) GT &  (d)  CPFP  & (e) A2dele  & (f) Ours \\
		\end{tabular}
	\end{center}
	\caption{Two motivating examples of SOD. (a)-(c) represent the input images, the corresponding depth maps, and the ground truth (GT), respectively. (d) and (e) are the results of state-of-the-art RGB-D SOD detectors  CPFP (\textbf{CVPR'19})  \cite{CPFP} and A2dele (\textbf{CVPR'20}) \cite{A2dele}, respectively. (f) are our results. \emph{Compared with the latest  CPFP and A2dele, our method can yield more complete, sharp, and edge-preserving saliency detection results by effectively intergrating cross-modality complementaries and adaptively selecting saliency-related features.}}
	\label{fig:example}
\end{figure}

Depth maps provide useful cues such as depth of field, shape, and boundary to complement RGB images for SOD \cite{PCFN,TAN,MMCI,DMRA,A2dele,CPFP}.
However, depth maps are inherently noisy and the cues provided can be inconsistent or misaligned with the RGB modality.
The issues make designing an RGB-D algorithm challenging. Contemporary RGB-D SOD detectors, CPFP \cite{CPFP} (Fig.~\ref{fig:example}(d)) and A2dele \cite{A2dele} (Fig.~\ref{fig:example}(e)), could still miss salient objects due to cluttered backgrounds or yield incomplete or serrated boundaries of saliency maps.

In this work, we consider addressing the aforementioned problem through more careful investigation on the integration of cross-modality complementaries from RGB image and depth map as well as the selection of saliency-related features.
To this end, we present an effective network that achieves complete, sharp, and edge-preserving saliency detection, as shown in Fig.~\ref{fig:example}(f).

First, we propose a cross-modality feature modulation (cmFM) module that  enhances RGB feature representations by taking the corresponding depth features as prior.
This is in contrast to popular strategies that perform either input fusion \cite{Peng2014}, early fusion \cite{DSS}, or late fusion \cite{CTMF}, that crudely concatenate or add the multi-modality information.
The proposed modulation design enables effective integration of multi-modality information through feature transformation, distinctly models the inseparable cross-modality relations, and reduces the interference caused by the inherent inconsistency of multi-modality data.

Second, we devise an adaptive feature selection (AFS) module that highlights the importance of different channel features in self- and cross-modalities, while fusing multi-modality spatial features in a gated manner.
This is different from previous RGB-D SOD algorithms \cite{PCFN,TAN,DCFF,MMCI,ASIF-Net,CPFP} that treat channel features from different modalities equally and independently.
Relaxing such assumptions allows our method to adaptively select more saliency-related features and suppress the inferior ones from both spatial features and channel features. It also mitigates the negative influence of poorly captured depth maps.
Hence, our network equips additional flexibility in dealing with different information.
We also emphasize the saliency-related positions and edges by introducing a saliency-guided position-edge attention (sg-PEA) module, which collects its attention weights from the predicted saliency maps and saliency edge maps.

Our method is unique in that the feature modulation and attention mechanism are closely coupled in a coarse-to-fine manner.
Specifically, fusion is first performed by the cmFM module to provide rich features representations. Coordinated with our AFS module, saliency-related features are emphasized while redundant features are suppressed. The saliency-related features are further refined by the sg-PEA module. A careful design to place the cmFM, AFS, and sg-PEA modules allows the cross-modality complementarities to go through modulation, selection, and refinement in a coarse-to-fine fashion, providing our network with precise saliency features.  Coupled with a bottom-up inference, the precise saliency features enable us to perform more accurate and robust SOD.

\noindent\textbf{Contributions}. We present an effective approach for RGB-D SOD. Cross-modality complementarities are effectively integrated and saliency-related features are adaptively selected. This is made possible by designing a coarse-to-fine fusion that consists of 1) a cross-modality feature modulation module that enhances RGB feature representations by taking the corresponding depth features as prior, and 2) an adaptive feature selection module that progressively emphasizes the importance of channel features in self- and cross-modalities while fusing the significant multi-modality spatial features.
Our method consistently outperforms state-of-the-art SOD methods on six popular RGB-D SOD benchmarks.

\section{Related Work}

\noindent
\textbf{Salient Object Detection.} SOD methods range from bottom-up \cite{DSR,SMD,RCRR} to top-down models \cite{AFNet,Guan2018,DSS,PoolNet,BASNet,EGNet}.
In addition to the color appearance, depth maps can provide useful cues such as depth of field, shape, and boundary.
%Therefore, RGB-D SOD has attracted much attention, which considers the color and depth attributes jointly.
The depth map is implicitly used in the unsupervised methods \cite{crm2019tc,DCMC,ACSD,Niu2012,Peng2014,Song2017,Zhu2018}. Whereas for the supervised methods, the discriminative and complementary features are learned from RGB-D images \cite{PCFN,TAN,DCFF,MMCI,fan2020bbs,Fu2020JLDCF,CTMF,ASIF-Net,DMRA,DF,Zhang2020CVPR,Zhangmiao2020CVPR,CPFP}.
%For example, Chen \etal~\cite{PCFN} designed a progressively complementarity-aware fusion module to integrate the cross-modal features.
%Piao \etal~\cite{DMRA} designed a depth-induced multi-scale recurrent attention network. Zhao \etal~\cite{CPFP} proposed a fluid pyramid integration module to integrate the RGB feature and depth feature enhanced by contrast prior.
%
Our work differs from recent works \cite{fan2020bbs,Fu2020JLDCF,DMRA,A2dele,Zhang2020CVPR,Zhangmiao2020CVPR,CPFP}, mainly in two aspects: 1) we use depth features as prior to learn optimal affine transformation parameters, which can flexibly modulate multi-level RGB features, and 2) we consider both self-modality and cross-modality channel features as well as multi-modality spatial features, thus effectively capturing relations among different modalities.

\noindent
\textbf{Feature Modulation.}
Inspired by FiLM \cite{FiLM} that first applies linear feature modulation for visual reasoning, feature modulation has been used in few-shot learning \cite{TADAM} and image super-resolution \cite{SR2018}.
%Boris \etal~\cite{TADAM} predicted the element-wise scale and shift vectors for each convolutional layer in the feature extractor for few-shot learning. Wang \etal~\cite{SR2018} proposed a spatial feature transform layer that modulates features of intermediate layers conditioned on the semantic segmentation probability maps for image super-resolution.
In our studies, we modulate the multi-level feature representations conditioned on the corresponding depth features. Besides, we design the cross-modality feature modulation in a pixel-wise manner, which provides elaborate and fine-grained control to the features.

\noindent
\textbf{Attention Mechanism.}
Attention mechanism is increasingly applied in diverse forms such as spatial attention \cite{Spattention}, dual-attention \cite{DualAttention}, self-attention  \cite{Attention17}, multi-level attention \cite{MultiAttention}, and channel attention \cite{ChannelAttention}.
In contrast, we employ the attention mechanism in our adaptive feature selection module, which explores the interdependencies of channel features in the self- and cross-modalities while fusing the significant multi-modality spatial features in a gated manner.

\section{Our Method}
We first present an overview of our network architecture. Then, we describe the key components including the cross-modality feature modulation module, adaptive feature selection module, and saliency-guided position-edge attention module. At last, we introduce the loss functions.

\subsection{Overview of Network Architecture }

\begin{figure*}[!t]
	\centering
	\includegraphics[width=0.98\textwidth]{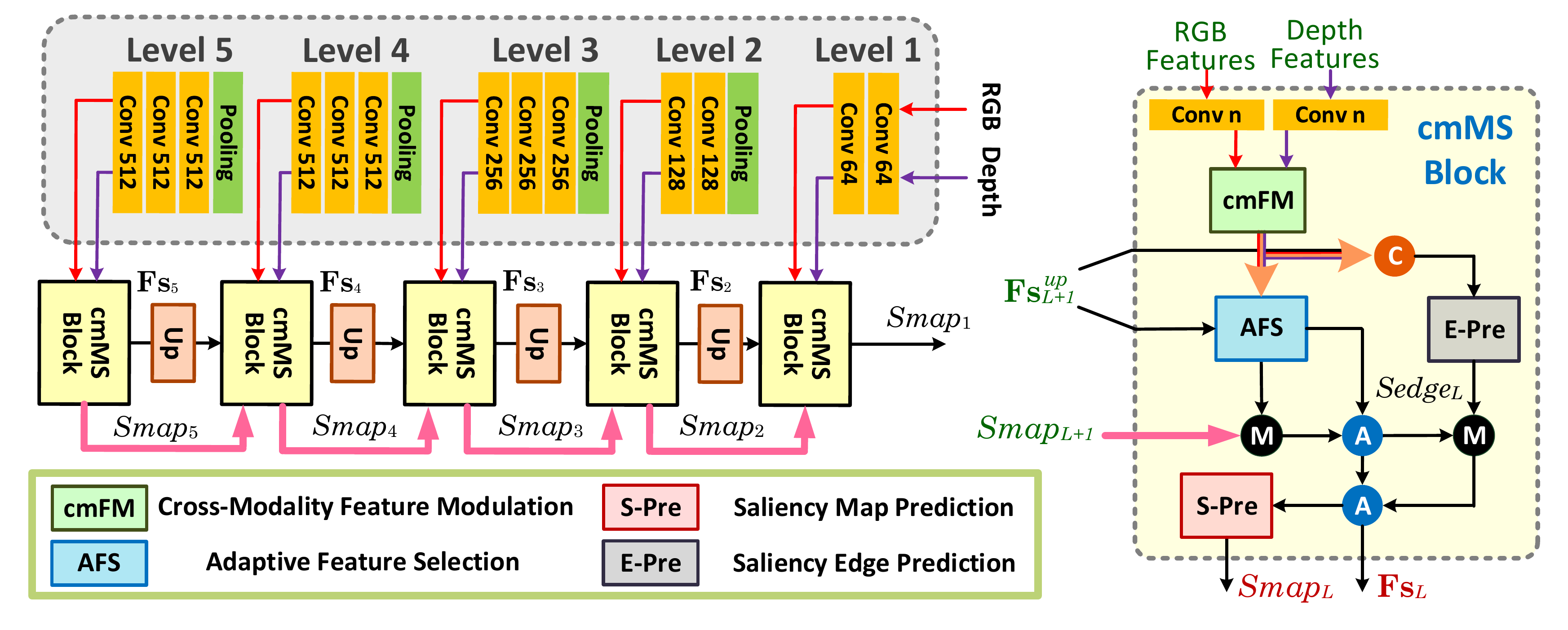}
	\caption{\textbf{Overview of our network architecture}. The inputs are the RGB image and its depth map. The cmMS block consists of a cmFM module, an AFS module, and an sg-PEA module. Here, the sg-PEA module further contains an S-Pre unit and an E-Pre unit. `Conv n' represents the convolutional layer that outputs $n$ feature maps, where $n$ is the half number of input feature maps.  `A', `M', and `C' represent element-wise addition, element-wise multiplication, and concatenation along with the channel dimension, respectively. `Up' represents the up-sampling block. Pink line indicates 2$\times$ linear interpolation. $\mathbf{Fs}$ represent the refined features after the cmMS block while  $\mathbf{Fs}^{up}$ are the up-sampled $\mathbf{Fs}$ by the `Up' block.  In this figure, each convolutional layer is followed by the ReLU activation. Our network finally produces five saliency maps ($Smap_{L}$) and five saliency edge maps ($Sedge_{L}$) with the resolutions, ranging from 14$\times$14 to 224$\times$224 by a scale of 2. $L$ indicates the level. We treat $Smap_{1}$ as the final result.}
	\label{framework}
\end{figure*}

The overview of our network architecture is illustrated in Fig.~\ref{framework}.
After the top-down features extraction from VGG-16 backbone \cite{VGG},
the multi-level RGB features and depth features are fed to a convolutional layer for halving the number of feature maps, respectively. Then, the dimension reduced RGB-D features are forwarded to the corresponding cmMS block. In each cmMS block, the RGB-D features go through cmFM module, AFS module, and sg-PEA module for feature modulation, selection, and refinement, respectively.
Specifically, we introduce modulated features by using our proposed cross-modality feature modulation (cmFM) module. The purpose of cmFM module is to effectively integrate the cross-modality complementarities in a flexible and trainable fashion.
After that, RGB features, depth features, modulated features, and  up-sampled features from the higher level (if any) are independently forwarded to our proposed adaptive feature selection (AFS) module for selectively emphasizing the informative channel features and fusing the significant spatial features. The AFS module models the relations between different levels and accelerates task-oriented feature integration.
Meanwhile, the concatenation of RGB features, depth features, modulated features, and up-sampled features (if any) is applied to predict the saliency edge map via a saliency edge prediction (E-Pre) unit.
Then, with the saliency map up-sampled from the higher level (if any) and saliency edge map, we highlight the saliency position and edge regions of the features after the AFS module.  After that, we predict the saliency map in the current level via a saliency map prediction (S-Pre) unit by using the refined features.
At last,  in the bottom-up inference, we progressively integrate and highlight multi-level features to predict the fine-scaled saliency map (\ie, the $Smap_{1}$ in Fig.~\ref{framework}).
We adopt 3$\times$3  kernels for all convolutional layers in our network, except the cmFM module that employs the multi-scale convolutions to enlarge receptive field.

\subsection{Cross-modality Feature Modulation (cmFM)}
\begin{figure*}[!ht]
	\centering
	\includegraphics[width=11cm,height=3.7cm]{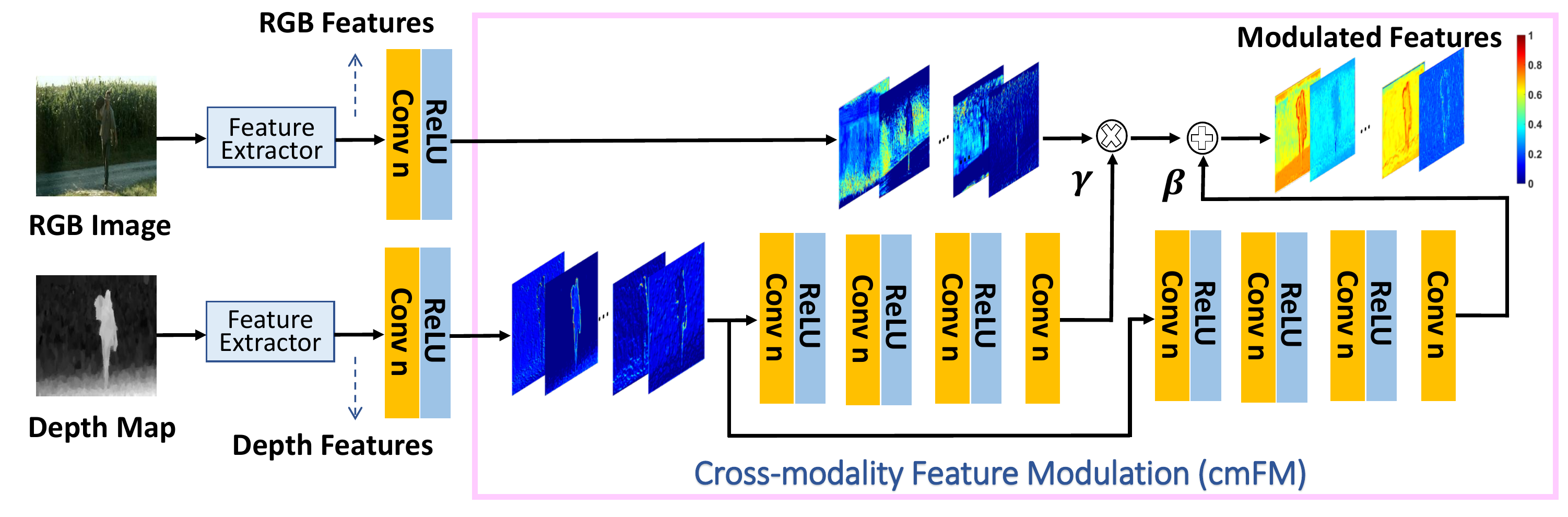}
	\caption{\textbf{The proposed cmFM module}. For the estimation of both  $\bm{\gamma}$ and $\bm{\beta}$, the kernels of convolutional layers are 7$\times$7, 5$\times$5, 3$\times$3, and 3$\times$3. The feature extractor represents VGG-16 backbone. The feature maps are illustrated as heatmaps.}
	\label{cmFM}
\end{figure*}

Inspired by the unsupervised RGB-D SOD algorithms \cite{DCMC,LBE}  which take the depth map as prior information to enrich the saliency cues, we propose a cmFM module conditioned on the depth features. The cmFM module learns pixel-wise affine transformation parameters from the conditioning depth features then modulates the corresponding RGB feature representations in each level of our network. The detailed cmFM module is illustrated in Fig. \ref{cmFM}.

Given the dimension halved RGB features $\mathbf{F}_{L}^{rgb}$ $\in$
$\mathbb{R}^{N\times H\times W}$ and depth features $\mathbf{F}_{L}^{depth}$ $\in$ $\mathbb{R}^{N\times H\times W}$,  the cmFM module learns a mapping function $\mathcal{M}$ conditioned on the depth features to yield a set of affine transformation parameters ($\bm{\gamma}_{L}$,$\bm{\beta}_{L}$) $\in$
$\mathbb{R}^{N\times H\times W}$. Here, $N$ is the number of feature maps; $H$ and $W$ are the height and width of the feature maps, respectively. It can be expressed as:
\begin{equation}
\label{equ_1}
(\bm{\gamma}_{L},\bm{\beta}_{L}) =\mathcal{M}(\mathbf{F}_{L}^{depth}),
\end{equation}
where the superscript indicates the modality while the subscript represents the level.  The mapping function $\mathcal{M}$ is built on two parallel stacked convolutional layers as shown in Fig. \ref{cmFM}.
With the estimated affine transformation parameters ($\bm{\gamma}_{L}$,$\bm{\beta}_{L}$), we conduct pixel-wise scaling and shifting on the RGB feature representations, which can be expressed as:
\begin{equation}
\label{equ_2}
\mathbf{F}_{L}^{mod}=\mathbf{F}_{L}^{rgb} \otimes \bm{\gamma}_{L}\oplus \bm{\beta}_{L},
\end{equation}
where $\mathbf{F}_{L}^{mod}$ represent the modulated features; $\otimes$ and $\oplus$ indicate the element-wise multiplication and element-wise addition, respectively. As shown in Fig. \ref{cmFM}, the cluttered backgrounds of RGB features become clear and the salient object is highlighted with the modulation of depth features.

\subsection{Adaptive Feature Selection (AFS)}

To make our network focus more on informative features, we propose an AFS module to progressively re-scale channel-wise features. Simultaneously, the AFS module fuses significant spatial features of multi-modalities. To be specific, we first explore the interdependencies of channel features in the self-modality, then further determine the relevance in the cross-modality. After squeezing by a convolutional layer that reduces the redundant features, we achieve the channel attention-on-channel attention features.  Such a self-modality and cross-modality channel attention mechanism can model relations of the channel features among different modalities well and adaptively select the informative channel features. The advantages of our channel attention-on-channel attention than the conventional channel attention are verified in the ablation studies.

\begin{figure*}[!t]
	\centering
	\includegraphics[width=0.92\textwidth]{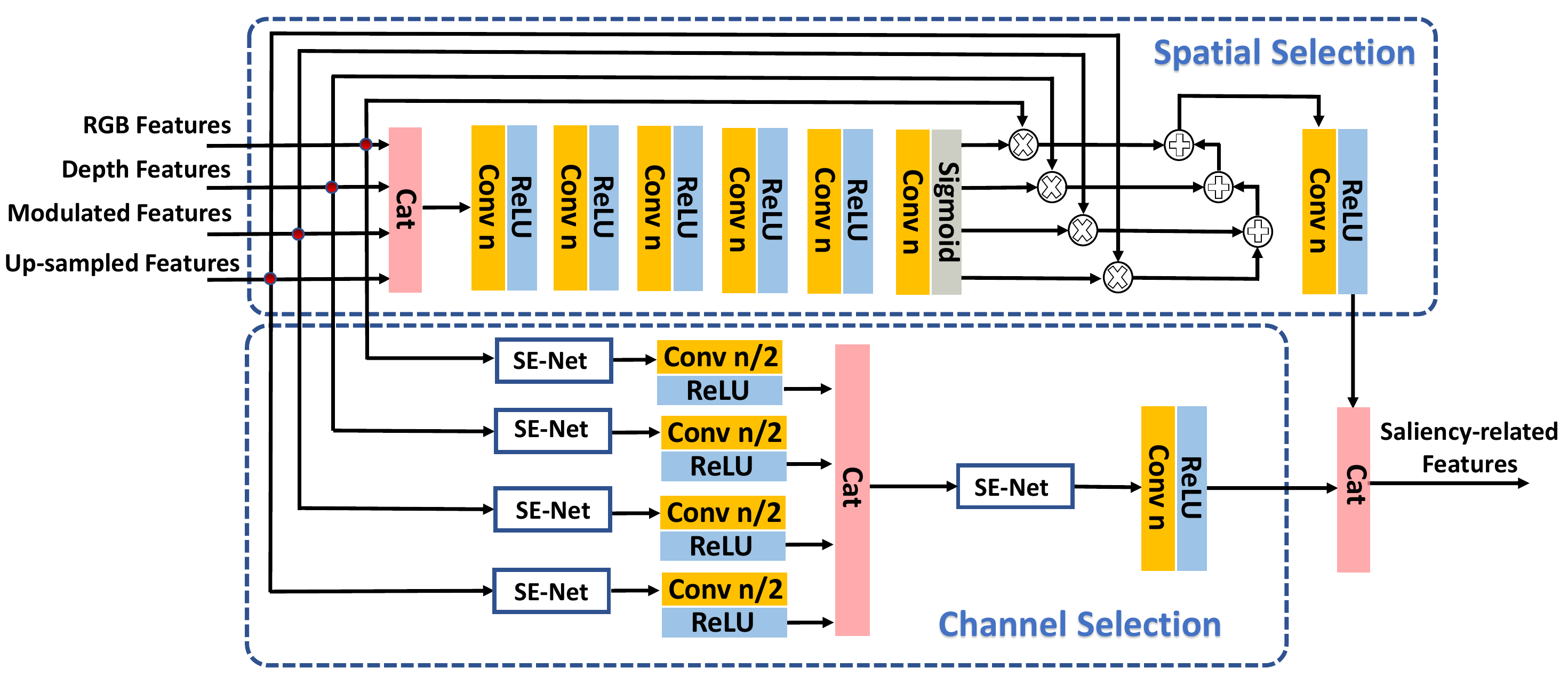}
	\caption{\textbf{The detail of AFS module}. `Cat' represents the concatenation operation. `SE-Net' is the squeeze-and-excitation network.}
	\label{CA-on-CA}
\end{figure*}

We simultaneously fuse the multi-modality features to achieve the enhanced feature representations based on a gated spatial fusion mechanism, where the pixel-wise confidence map for each input feature is calculated. In this way, the significant multi-modality spatial features are preserved. As a result, we achieve saliency-related features and filter out irrelevant or misleading features from both  spatial and channel aspects. The detail of AFS module is shown in Fig. \ref{CA-on-CA}.

Given the features ($\mathbf{F}_{L}^{rgb}$, $\mathbf{F}_{L}^{depth}$, $\mathbf{F}_{L}^{mod}$, $\mathbf{Fs}_{L+1}^{up}$), we first perform global average pooling on each set of features separately, leading to a channel descriptor $\mathbf{z}\in\mathbb{R}^{N\times 1}$ for each one, which is an embedded global distribution of channel-wise feature responses. $\mathbf{Fs}_{L+1}^{up}$ indicate 2$\times$ up-sampled features from the $L$+1 level by using the `Up' block that consists of one 2$\times$ linear interpolation followed by two convolutional layers, where each convolutional layer is followed by the ReLU activation and outputs $n$ feature maps. The $k$-th entry of $\mathbf{z}$ is expressed as:
\begin{equation}
\label{equ_3}
z_{k} =\frac{1}{H\times W} \sum_{i}^{H} \sum_{j}^{W} \mathbf{F}_{k}(i,j),
\end{equation}
where $k\in[1,N]$. Then, a self-gating mechanism is used to fully capture channel-wise dependencies $\mathbf{s}\in\mathbb{R}^{N\times 1}$:
\begin{equation}
\label{equ_4}
\mathbf{s}=\mathscr{\sigma}(\mathbf{W}_{2}\ast(\mathscr{\delta}(\mathbf{W}_{1}\ast\mathbf{z}))),
\end{equation}
where $\mathscr{\sigma}(\cdot)$ represents the Sigmoid activation,  $\mathscr{\delta}(\cdot)$ represents the ReLU activation, $\ast$ denotes the convolution operation, and $\mathbf{W}_{1}$ and $\mathbf{W}_{2}$ are the weights of two fully-connected layers with their numbers of output channels being $\frac{N}{16}$ and $N$, respectively. At last, these weights are applied to each set of input features $\mathbf{F}$ to generate re-scaled features $\mathbf{U}$ $\in$ $\mathbb{R}^{N\times H\times W}$: $\mathbf{U} =\mathbf{F}\otimes \mathbf{s}$.
This processing is mathematically expressed as an $SE$ mapping function in this paper and  can also be implemented by the squeeze-and-excitation network \cite{ChannelAtt}. However, the highlighted channel features may become relatively useless among all channel attention results from multi-modalities.

To emphasize the informative channel features, we first halve the number of feature maps in each channel attention result by a convolutional layer, then concatenate them: $\mathbf{V}_{L} =Cat\{\mathbf{U}_{L}^{rgb}, \mathbf{U}_{L}^{depth},\mathbf{U}_{L}^{mod}, \mathbf{Us}_{L+1}^{up}\}$.
After that, we further explore the interdependencies of channel features by $\mathbf{Y}_{L} = SE(\mathbf{V}_{L})$.
We finally squeeze the number of channel features by a convolutional layer and achieve the results of channel attention-on-channel attention  $\mathbf{Y}_{L}^{caca}$.

\begin{figure*}[!t]
	\centering
	\includegraphics[width=12cm,height=1.8cm]{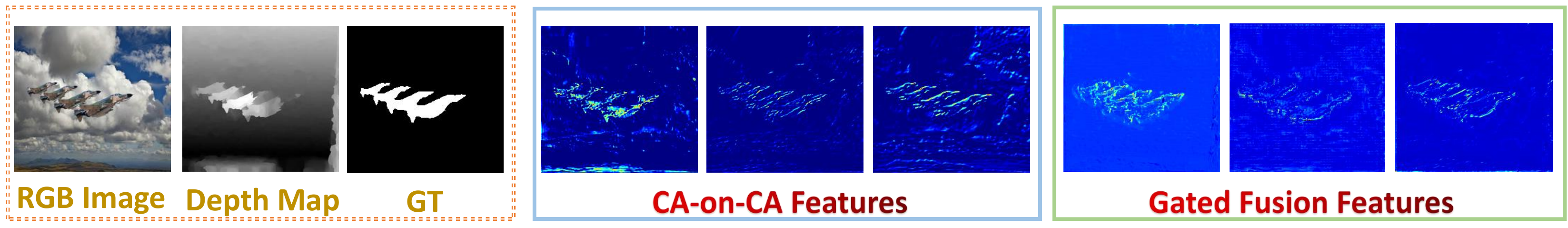}
	\caption{\textbf{Visual results of the intermediate features in our AFS module}. `CA-on-CA Features' indicates the features after our channel selection while `Gated Fusion Features' represents the features after our spatial selection.}
	\label{Feature_map}
\end{figure*}

Meanwhile, we fuse the multi-modality input features to achieve enhanced spatial feature representations.
First, the input features are concatenated $\mathbf{F}_{L}^{cat}=Cat\{\mathbf{F}_{L}^{rgb}, \mathbf{F}_{L}^{depth}, \mathbf{F}_{L}^{mod}, \mathbf{Fs}_{L+1}^{up}\}$, and fed to a plain CNN network (indicated as $\mathcal{G}$) to estimate their pixel-wise confidence maps:
\begin{equation}
\label{equ_8}
(\mathbf{C}_{L}^{rgb}, \mathbf{C}_{L}^{depth}, \mathbf{C}_{L}^{mod}, \mathbf{C}_{L+1}^{up}) = \mathcal{G}(\mathbf{F}_{L}^{cat}),
\end{equation}
where $\mathbf{C}_{L}^{rgb}$, $\mathbf{C}_{L}^{depth}$, $\mathbf{C}_{L}^{mod}$, and $\mathbf{C}_{L+1}^{up}$ $\in$ $\mathbb{R}^{N\times H\times W}$ represent the confidence maps. The $\mathcal{G}$ is built on six stacked convolutional layers as shown in Fig. \ref{CA-on-CA}. The achieved features in the level $L$ can be expressed as:
\begin{equation}
\label{equ_9}
\mathbf{F}_{L}^{gated}=\mathbf{F}_{L}^{rgb}\otimes \mathbf{C}_{L}^{rgb} \oplus \mathbf{F}_{L}^{depth} \otimes\mathbf{C}_{L}^{depth}\oplus\mathbf{F}_{L}^{mod} \otimes\mathbf{C}_{L}^{mod}\oplus \mathbf{Fs}_{L+1}^{up} \otimes\mathbf{C}_{L+1}^{up}
\end{equation}
Then, we pass these features to a convolutional layer and achieve the gated fusion features $\mathbf{F}_{L}^{gated'}$. At last, we combine the enhanced spatial feature representations with the enhanced channel feature representations by:
\begin{equation}
\label{equ_10}
\mathbf{F}_{L}^{AFS}=Cat\{\mathbf{F}_{L}^{gated'}, \mathbf{Y}_{L}^{caca}\},
\end{equation}
where the final results $\mathbf{F}_{L}^{AFS}$ enjoy the most informative features towards saliency detection, called saliency-related features in this paper. The visual examples are presented in Fig. \ref{Feature_map}. As shown, the saliency-related spatial features and channel features are preserved and highlighted.

\subsection{Saliency-Guided Position-Edge Attention (sg-PEA)}
After selecting the saliency-related features, we also encourage the network to focus on those positions and edges most essential to the nature of salient objects. The benefits are illustrated as follows:
1) the saliency position attention can better locate the salient objects and accelerate the network convergence, and 2) the saliency edge attention can alleviate the problem of edge blur caused by the repeated pooling operations, which is vital for the pixel-wise saliency prediction.

\begin{figure*}[!t]
	\centering
	\includegraphics[width=12cm,height=3cm]{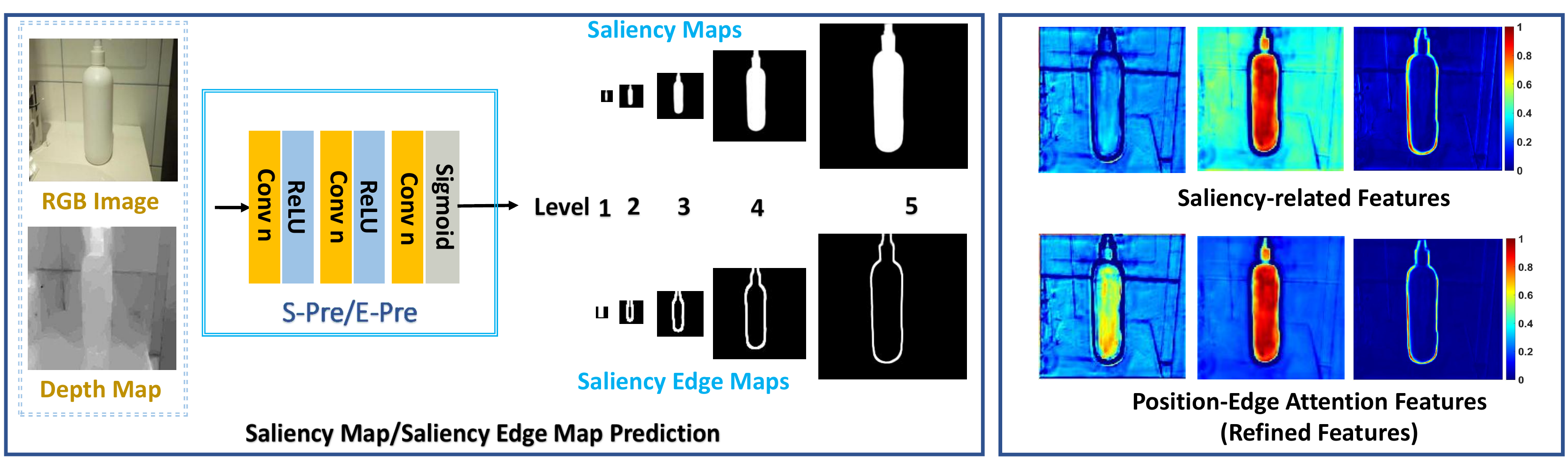}
	\caption{\textbf{Visual results of sg-PEA module}. Left panel shows the structure of S-Pre/E-Pre unit, and the predicted saliency maps and saliency edge maps in different levels. Right panel shows the intermediate features before and after the sg-PEA module. After the sg-PEA module, the background of features are suppressed, and the edge and position details are assigned more focuses. }
	\label{PEA}
\end{figure*}

To the end, we propose a saliency-guided position-edge attention (sg-PEA) module to  locate and sharpen salient objects. The sg-PEA module further includes a saliency map prediction  (S-Pre) unit and a saliency edge prediction (E-Pre) unit as shown in Fig. \ref{framework}. The details are provided in Fig. \ref{PEA}, where  S-Pre unit and E-Pre unit share the same structure, but different weights.

\noindent
\textbf{Position Attention.} We employ the up-sampled saliency map from the higher level as the attention weights. Here, the up-sampling is implemented by the simple 2$\times$ linear interpolation. In our method, the saliency map is predicted by the S-Pre unit in each level in a supervised learning manner.
The benefits of such a side supervision manner lie in four aspects: 1) the convolutional layers in each level have explicit objective towards saliency detection, 2) the side supervision can accelerate gradient back-propagation, 3) the predicted saliency map works as a guidance and can steer the convolutional layers of lower level to focus more on saliency positions in a low-computational manner, and 4) the multiple side outputs can provide diverse choices based on accuracy and inference speed. We provide more analysis on the side outputs in the supplementary material.

To be specific, with the saliency-related features $\mathbf{F}_{L}^{AFS}$ and the up-sampled saliency map $Smap_{L+1}^{up}$, the position attention results $\mathbf{F}_{L}^{poa}$ can be expressed as:
\begin{equation}
\label{equ_11}
\mathbf{F}_{L}^{poa} =\mathbf{F}_{L}^{AFS} \oplus \mathbf{F}_{L}^{AFS} \otimes Smap_{L+1}^{up}
\end{equation}
In contrast to treating all positions of saliency features equally, the position attention can quickly and efficiently employ the saliency property of higher level and enhance the  saliency representations of the current level. To avoid gradient diffusion induced by successive attention (the values of feature maps are close to zero), we adopt an identical mapping manner as shown in Eq. \eqref{equ_11}.

\noindent
\textbf{Edge Attention.}
To obtain the edge attention weights, we first concatenate the RGB-D features, the modulated features, and up-sampled features, then forward them to the E-Pre unit to predict the saliency edge map in each level. The saliency edge maps, also estimated by supervised learning, can be used to emphasize the salient edges of the features by simple element-wise multiplication. For level $L$, the output features of edge attention can be expressed as:
\begin{equation}
\label{equ_12}
\mathbf{Fs}_{L} =\mathbf{F}_{L}^{poa}\oplus \mathbf{F}_{L}^{poa} \otimes Sedge_{L},
\end{equation}
where $Sedge_{L}$ is the predicted saliency edge map in the level $L$. We call $\mathbf{Fs}_{L}$ as the refined features. At last, with the refined features, the  final result (\ie, $Smap_{1}$)  with the same size as the input RGB image can be achieved in a bottom-up manner. In Fig. \ref{PEA}, we present the changes of features before and after sg-PEA module. As shown, the features increasingly focus on the saliency position and edge details, while the cluttered backgrounds are concurrently reduced.

\subsection{Loss Function}
We employ the standard cross-entropy (SCE) loss \cite{cross-entropy} to  jointly optimize our network for the saliency prediction and saliency edge prediction:
\begin{equation}
\label{equ_13}
Loss=\sum_{i=1}^{L} (\lambda_{i}SCE_{i}^{SPre}+\eta_{i}SCE_{i}^{EPre}),
\end{equation}
where $L$ indicates the level, $SCE_{i}^{SPre}$ and $SCE_{i}^{EPre}$ represent the losses for predicting the saliency map and saliency edge map in the level $i$, respectively. $\lambda$ and $\eta$ are the corresponding weights.

\section{Experiments}\label{er}
\subsection{Benchmark Datasets and Evaluation Metrics}

We conduct experiments on six popular RGB-D SOD datasets, including \textbf{NJUD} \cite{ACSD} (1985 RGB-D images), \textbf{NLPR} \cite{Peng2014} (1000 RGB-D images), \textbf{STEREO} \cite{Niu2012} (797 RGB-D images), \textbf{LFSD} \cite{LFSD} (100 RGB-D images), \textbf{SSD} \cite{SSD} (80 RGB-D images), and \textbf{DUT} \cite{DMRA} (1200 RGB-D images).
For quantitative evaluations, Precision-Recall (P-R) curve,  F-measure \cite{Fmeasure}, MAE score \cite{RERVIEW}, and S-measure \cite{S-measure} are employed. P-R curve depicts the different combinations of precision and recall scores; the closer the P-R curve is to (1,1), the better the performance of the method. F-measure is the weighted harmonic mean of precision and recall; it is a comprehensive measurement, with a larger value indicating a better performance. MAE score measures the difference between the continuous saliency map and ground truth; a smaller value indicates a smaller gap hence better. S-measure  calculates the structural similarity between the saliency map and ground truth; a larger value indicates a better performance. Additionally, we compare the model sizes of different methods in the supplementary material.

\subsection{Implementation Details}
We adopt the same training, validation, and testing sets as described in \cite{DMRA,A2dele}. The ground truth of saliency edge map prediction is obtained by using the Canny edge detector on the saliency mask. We implement our network with TensorFlow on a PC with an Nvidia Tesla V100 GPU. During training, the batch size is set to 4, the filter weights of each layer are initialized by Gaussian distribution, and the bias is initialized as a constant. We use ADAM and fix the learning rate to $1e^{-4}$. The weight $\lambda_{1}$ for predicting the final saliency map is set to 1.2 while other weights are set to 1 in Eq. \eqref{equ_13}. For a pair of RGB-D images of size 224$\times$224, the average runtime of our method is 0.037s on the aforementioned PC.

\subsection{Comparisons with State-of-the-art Methods}
We compare our method with 12 state-of-the-art learning-based SOD methods, including two latest RGB-induced SOD methods (\ie, PoolNet \cite{PoolNet} and EGNet \cite{EGNet}), and ten RGB-D SOD methods (\ie, DF \cite{DF}, CTMF \cite{CTMF}, MMCI \cite{MMCI}, PCFN \cite{PCFN}, TAN \cite{TAN}, CPFP \cite{CPFP}, DCFF \cite{DCFF}, DMRA \cite{DMRA}, ASIF-Net \cite{ASIF-Net}, and A2dele \cite{A2dele}).

\begin{figure*}[!t]
	\centering
	\includegraphics[width=0.95\textwidth]{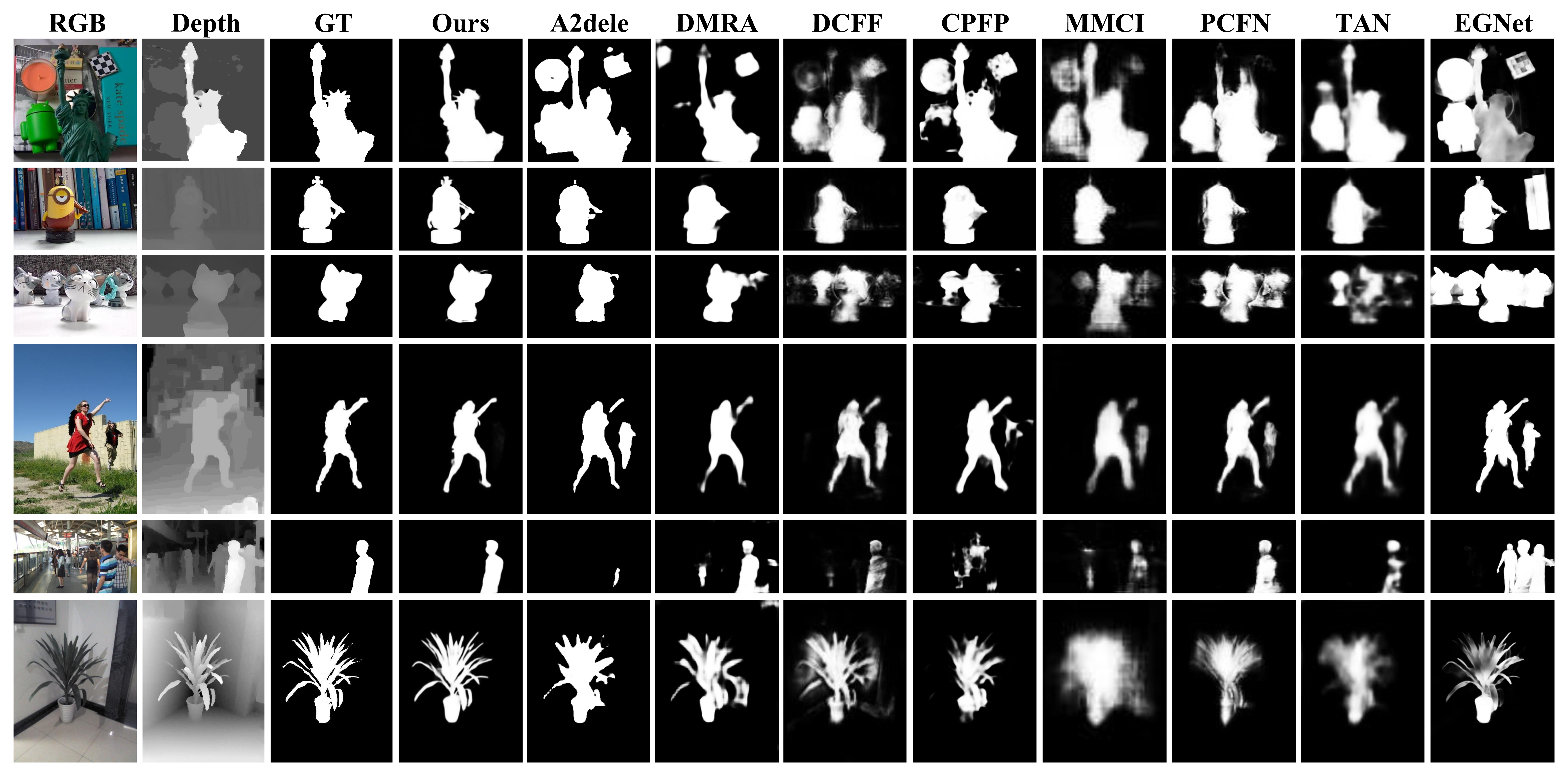}
	\caption{Visual examples of different methods.}
	\label{visual}
\end{figure*}

\begin{figure*}[!t]
	\centering
	\includegraphics[width=0.95\textwidth]{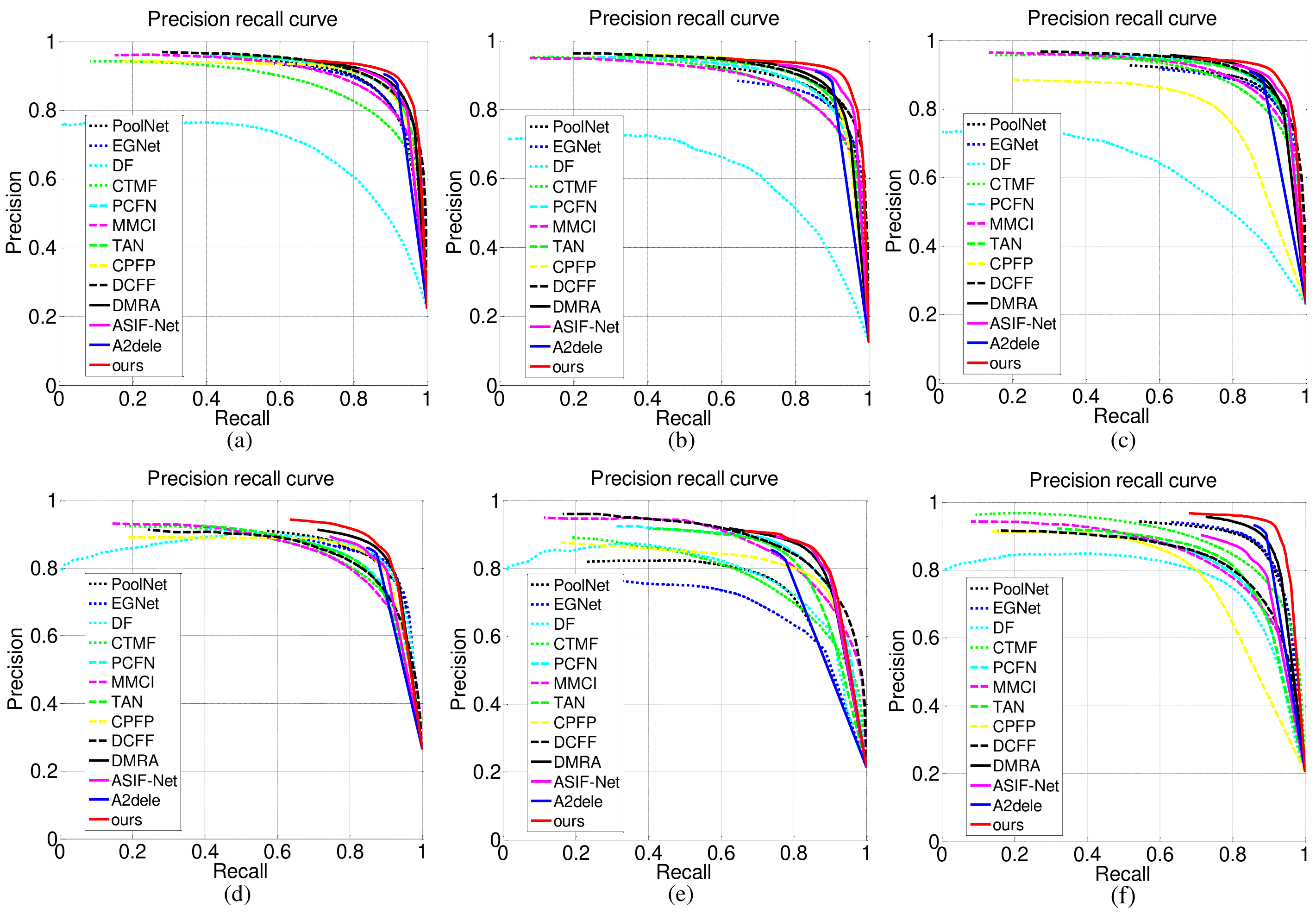}
	\caption{P-R curves of different methods on the testing datasets. (a)-(f) correspond to STEREO, NLPR-Test, NJUD-Test, LFSD, SSD, and DUT-Test datasets. }
	\label{PR}
\end{figure*}

Visual comparisons are shown in Fig. \ref{visual}. Our method achieves more competitive performance than the compared methods.
\textbf{First}, the salient objects in our results are more complete and accurate, and the object boundaries are sharper. In the first image, only our method can accurately and completely detect the salient toy in front, while the competing methods incorrectly reserve the background regions (\eg, Android doll and checkerboard). In the fourth image that comes with an unsatisfactory depth map, our method can still accurately locate salient target with a complete structure and clear boundaries.
\textbf{Second}, our method preserves more details in the saliency result. In the sixth image, more details of plant leaves are better conserved.
\textbf{Third}, our method can address some challenging cases, such as a complex background and small object. In the third image, the cat dolls in the back row are successfully suppressed by our method, the detected salient boundaries are sharper, and the structure is more complete. In the fifth image illustrating a case of complex background, our method can still completely detect a small salient object (\ie, the human).
%, while other methods fail to suppress the interference of background noise.

\begin{table*}[t]
	\renewcommand\arraystretch{0.85}
	\caption{Quantitative comparisons on six testing datasets. The bold numbers are performance of our method, also the best across all datasets}
	\begin{center}
		\begin{tabular}{|c|c|c|c||c|c|c||c|c|c|}
			\hline
			\multirow{2}{*}{} & \multicolumn{3}{c||}{STEREO Dataset} & \multicolumn{3}{c||}{NLPR-Test Dataset} & \multicolumn{3}{c|}{NJUD-Test Dataset} \\[0.5ex]
			\cline{2-10}
			\cline{2-10}
			& $F_{\beta}\uparrow$ & MAE $\downarrow$ & $S_m \uparrow$ & $F_{\beta}\uparrow$ & MAE $\downarrow$ & $S_m \uparrow$ & $F_{\beta}\uparrow$ & MAE $\downarrow$ & $S_m \uparrow$  \\
			\hline\hline
			
			PoolNet \cite{PoolNet} & 0.8757 & 0.0655 & 0.8359 & 0.8627 & 0.0448 & 0.8573 & 0.8740 & 0.0676 & 0.8600  \\
			
			EGNet \cite{EGNet} & 0.8717 & 0.0671 & 0.8363 & 0.8452 & 0.0504 & 0.8497 & 0.8667 & 0.0704 & 0.8562 \\
			
			DF \cite{DF}  & 0.6961 & 0.1738 & 0.6279 & 0.6480 & 0.1079 & 0.6710 & 0.6355 & 0.1987 & 0.5930  \\
			
			CTMF \cite{CTMF}  & 0.8265 & 0.1023 & 0.8230 & 0.8407 & 0.0561 & 0.8549 & 0.8572 & 0.0847 & 0.8493  \\
			
			PCFN \cite{PCFN}  & 0.8838 & 0.0606 & 0.8722 & 0.8635 & 0.0437 & 0.8592 & 0.8875 & 0.0592 & 0.8768  \\
			
			MMCI \cite{MMCI}  & 0.8610 & 0.0796 & 0.8504 & 0.8412 & 0.0591 & 0.8524 & 0.8684 & 0.0789 & 0.8588  \\
			
			TAN \cite{TAN} & 0.8865 & 0.0591 & 0.8701 & 0.8765 & 0.0410 & 0.8736 & 0.8882 & 0.0605 & 0.8785  \\
			
			CPFP \cite{CPFP}  & 0.8856 & 0.0537 & 0.8702 & 0.8878 & 0.0359 & 0.8760 & 0.7994 & 0.0794 & 0.7984  \\
			
			DCFF \cite{DCFF} & 0.8867 & 0.0638 & 0.8706 & 0.8779 & 0.0439 & 0.8695 & 0.8910 & 0.0646 & 0.8774  \\
			
			DMRA \cite{DMRA} & 0.8953 & 0.0474 & 0.8778 & 0.8870 & 0.0339 & 0.8646 & 0.9003 & 0.0529 & 0.8804  \\
			
			ASIF-Net \cite{ASIF-Net} & 0.8939 & 0.0493 & 0.8686 & 0.9002 & 0.0298 & 0.8844 & 0.9007 & 0.0471 & 0.8887  \\
			
			A2dele \cite{A2dele} & 0.8997 & 0.0431 & 0.8713 & 0.8976 & 0.0285 & 0.8770 & 0.8939 & 0.0510 & 0.8704  \\

			ours & \textbf{0.9084} & \textbf{0.0422} & \textbf{0.8895} & \textbf{0.9137} & \textbf{0.0273} & \textbf{0.8999} & \textbf{0.9149} & \textbf{0.0442} & \textbf{0.9040}  \\
			\hline
			\hline
			\multirow{2}{*}{} & \multicolumn{3}{c||}{LFSD Dataset} & \multicolumn{3}{c||}{SSD Dataset} & \multicolumn{3}{c|}{DUT-Test Dataset} \\[0.5ex]
			\cline{2-10}
			& $F_{\beta}\uparrow$ & MAE $\downarrow$ & $S_m \uparrow$ & $F_{\beta}\uparrow$ & MAE $\downarrow$ & $S_m \uparrow$ & $F_{\beta}\uparrow$ & MAE $\downarrow$ & $S_m \uparrow$  \\
			\hline\hline
			
			PoolNet \cite{PoolNet} & 0.8474 & 0.0945 & 0.8217 & 0.7644 & 0.1099 & 0.7491 & 0.8828 & 0.0669 & 0.8392 \\

			EGNet \cite{EGNet} & 0.8445 & 0.0871 & 0.8300 & 0.7040 & 0.1351 & 0.7072 & 0.8876 & 0.0641 & 0.8439 \\
			
			DF \cite{DF}      & 0.8534 & 0.1424 & 0.7791 & 0.7631 & 0.1511 & 0.7422 & 0.7747 & 0.1455 & 0.7051 \\
			
			CTMF \cite{CTMF}  & 0.8147 & 0.1202 & 0.7883 & 0.7550 & 0.1003 & 0.7757 & 0.8417 & 0.0971 & 0.8226 \\
			
			PCFN \cite{PCFN}  & 0.8290 & 0.1118 & 0.7919 & 0.8447 & 0.0627 & 0.8427 & 0.8094 & 0.0999 & 0.7878 \\
			
			MMCI \cite{MMCI}  & 0.8128 & 0.1318 & 0.7793 & 0.8230 & 0.0820 & 0.8133 & 0.8044 & 0.1125 & 0.7818 \\
			
			TAN \cite{TAN}    & 0.8275 & 0.1108 & 0.7935 & 0.8350 & 0.0629 & 0.8393 & 0.8236 & 0.0926 & 0.7948 \\
			
			CPFP \cite{CPFP}  & 0.8495 & 0.0881 & 0.8200 & 0.8014 & 0.0818 & 0.8067 & 0.7866 & 0.0995 & 0.7335 \\
			
			DCFF \cite{DCFF}  & 0.8220 & 0.1191 & 0.7917 & 0.8388 & 0.0769 & 0.8316 & 0.8141 & 0.1014 & 0.7835 \\
			
			DMRA \cite{DMRA} & 0.8723 & 0.0754 & 0.8391 & 0.8579 & 0.0583 & 0.8569 & 0.9082 & 0.0477 & 0.8637 \\
			
			ASIF-Net \cite{ASIF-Net} & 0.8584 & 0.0896 & 0.8144 & 0.8633  & 0.0562 & 0.8566 & 0.8574  &0.0725  &  0.8141  \\
			
			A2dele \cite{A2dele} & 0.8577 & 0.0740 & 0.8306 & 0.8248 & 0.0691 & 0.8093 & 0.9145 & 0.0426 & 0.8611  \\
			
			ours  & \textbf{0.8882} & \textbf{0.0720} & \textbf{0.8465} & \textbf{0.8650} & \textbf{0.0524} & \textbf{0.8615} & \textbf{0.9328} & \textbf{0.0366} & \textbf{0.8853} \\
			\hline
		\end{tabular}
	\end{center}
	\label{num1}
\end{table*}

The P-R curves of different methods are shown in Fig. \ref{PR}. Our method (\ie, the red solid line) achieves the highest precision compared to other methods on all datasets. The numerical results are reported in Table \ref{num1}. Our method achieves the best quantitative results across all metrics, outperforming all competing methods.  Compared with the \textbf{second best method} on the NJUD-Test dataset, the percentage gain reaches $1.6\%$ for F-measure, $6.2\%$ for MAE score, and $1.7\%$ for S-measure. On the DUT-test dataset, the \textbf{minimum percentage gain} reaches $2.0\%$ for F-measure, $14.1\%$ for MAE score, and $2.5\%$ for S-measure. All these measures demonstrate the superiority and effectiveness of our method.

\begin{figure*}[!t]
	\centering
	\includegraphics[width=0.9\textwidth]{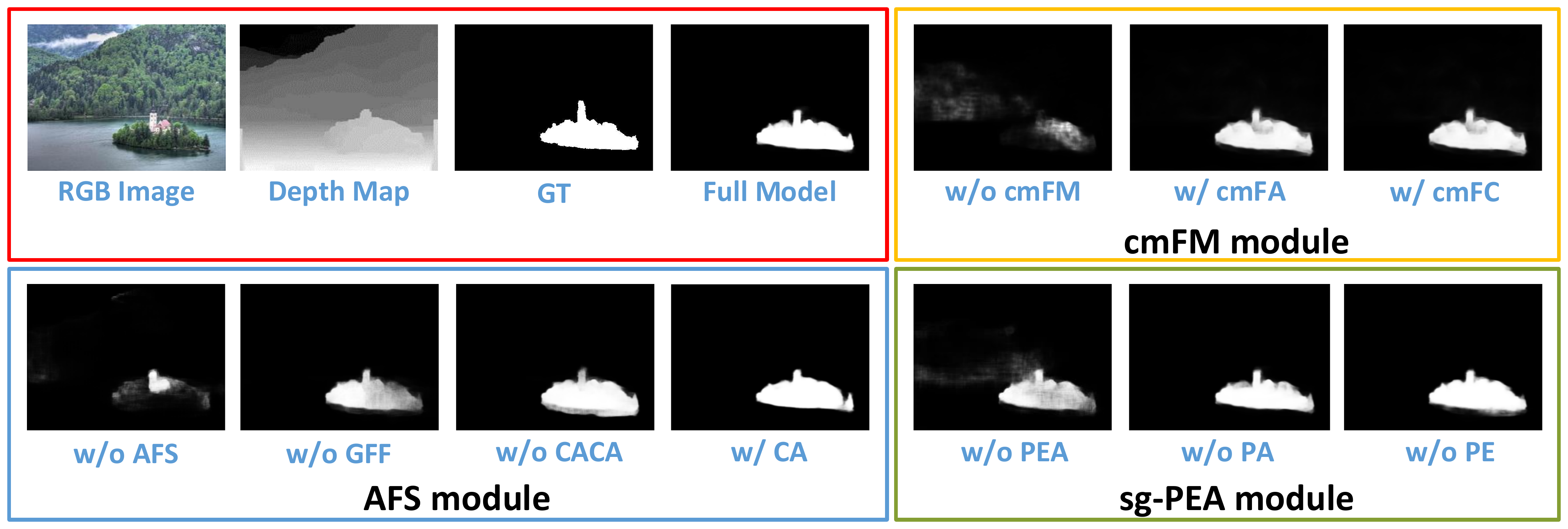}
	\caption{Visual comparison with different baselines. \textbf{(1)} The baseline \textit{w/o cmFM} represents our full model without the cmFM module (\ie, no modulated features); and the baselines \textit{w/ cmFA} and \textit{w/ cmFC} refer to that the cmFM module is replaced by the cmFA or cmFC module (\ie, the depth and RGB features are integrated by the element-wise addition or concatenation). \textbf{(2)} The baseline \textit{w/o AFS} represents our full model without the AFS module (\ie, the features after cmFM module are directly concatenated with the up-sampled saliency-related features); the baselines \textit{w/o GFF} and \textit{w/o CACA} correspond to removing the fused spatial features and the channel attention-on-channel attention features, respectively; and the baseline  \textit{w/ CA} refers to that the AFS module is replaced by the conventional channel attention module \cite{ChannelAtt}. \textbf{(3)} The baselines \textit{w/o PEA}, \textit{w/o PA}, and \textit{w/o PE} correspond to our full model without the sg-PEA module, the position attention unit, and the edge attention unit, respectively.}
	\label{ablation_results}
\end{figure*}

\subsection{Ablation Studies}

To verify the impact of our key modules, we conduct experiments on the STEREO dataset and DUT-Test dataset. The quantitative results are shown in Table \ref{num3}. An example of visual comparison is illustrated in Fig. \ref{ablation_results}.
% Our full model achieves the best quantitative performance across two testing datasets.

%
\noindent
\textbf{Cross-Modality Feature Modulation (cmFM).} We compare three variants: \textit{w/o cmFM}, \textit{w/ cmFA}, and \textit{w/ cmFC}. In Fig. \ref{ablation_results}, the baseline  \textit{w/o cmFM} cannot effectively detect the salient object while the baselines \textit{w/ cmFA} and \textit{w/ cmFC} achieve the similar detection result. The same quantitative trend also reflects in  Table \ref{num3}. Compared with the full model, the results indicate that the proposed cmFM module is important for improving the SOD performance. Besides, the simple addition and concatenation can only boost a little performance.

\noindent
\textbf{Adaptive Feature Selection (AFS).}
We compare with four baselines: \textit{w/o AFS}, \textit{w/o GFF}, \textit{w/o CACA}, and \textit{ w/ CA}.   Observing Fig. \ref{ablation_results} and Table \ref{num3}, we found that the performance of the baseline  \textit{w/o AFS} is obviously worse than the baselines \textit{w/o GFF}, \textit{w/o CACA}, and \textit{w/ CA}. The visual results reflect that the baseline \textit{w/o GFF} produces incomplete salient object while the baseline \textit{w/o CACA} yields the result with an unclear boundary.  Collectively, these results underscore the importance of progressive self-modality and cross-modality channel attention while fusing important spatial features of multi-modalities.

\noindent
\textbf{Saliency-guided Position-Edge Attention (sg-PEA).} We compare with three baselines: \textit{w/o PEA}, \textit{w/o PA}, and \textit{w/o EA}. In Fig. \ref{ablation_results}, the baseline \textit{w/o PEA} fails to highlight the position and edge of salient object. The baseline \textit{w/o PA} has a sharper boundary of partial complete object while the baseline \textit{w/o PE} shows a more complete object but unclear boundary. In contrast, our full model achieves better performance than these three baselines as presented in Table \ref{num3}.

In summary, the ablation studies demonstrate the effectiveness and advantages of the proposed three modules qualitatively and quantitatively. In addition, the ablation studies also demonstrate that careful feature modulation, selection, and refinement can effectively improve the performance of RGB-D SOD.

\begin{table*}[!t]
	\renewcommand\arraystretch{0.85}
	\caption{Quantitative comparisons of ablated models}
	\begin{center}
		\setlength{\tabcolsep}{1mm}{
			\begin{tabular}{|c|c|c|c|c||c|c|c|}
				\hline
				\multirow{2}{*}{Modules} & \multirow{2}{*}{Baselines} & \multicolumn{3}{c||}{STEREO Dataset} & \multicolumn{3}{c|}{DUT-Test Dataset} \\[0.5ex]
				\cline{3-8}
				&  & $F_{\beta}\uparrow$ & MAE $\downarrow$ & $S_m \uparrow$ & $F_{\beta}\uparrow$ & MAE $\downarrow$ & $S_m \uparrow$\\
				\hline\hline
				& \textbf{full model} &\textbf{0.9084} & \textbf{0.0422} & \textbf{0.8895} &\textbf{0.9328} & \textbf{0.0366} & \textbf{0.8853}  \\
				\hline
				\multirow{3}{*}{cmFM}&w/o cmFM & 0.8727 & 0.0722 & 0.8573 & 0.8968 &0.0616 & 0.8599 \\
				& w/ cmFA & 0.9020 & 0.0479 &0.8820 & 0.9237 & 0.0429 & 0.8771 \\
				& w/ cmFC & 0.8995 & 0.0480 & 0.8825 & 0.9221 &0.0617 & 0.8789 \\
				\hline
				\multirow{4}{*}{AFS}&w/o AFS & 0.8990 & 0.0546 & 0.8762 & 0.9165 & 0.0503 & 0.8666 \\
				& w/o GFF & 0.9012 & 0.0690 &0.8826 & 0.9212  & 0.0458 &0.8777 \\
				& w/o CACA &0.9017 &0.0517 &0.8797 & 0.9276 & 0.0470 & 0.8742 \\
				& w/ CA &0.9027  & 0.0503 &0.8780 & 0.9216 & 0.0468 & 0.8747 \\
				\hline
				\multirow{3}{*}{sg-PEA}&w/o PEA & 0.9057 & 0.0450 & 0.8854 & 0.9205 & 0.0427 & 0.8796 \\
				& w/o PA & 0.9064 & 0.0442 & 0.8857 & 0.9234 & 0.0409 & 0.8827 \\
				& w/o PE & 0.9065 & 0.0481 & 0.8862 & 0.9296 & 0.0385 & 0.8806 \\
				\hline
%				\multirow{4}{*}{Side Output}&level 2 & 0.9090 & 0.0448 &  0.8894 & 0.9322 & 0.0388  & 0.8887 \\
%                 & level 3 & 0.9057 &0.0501 & 0.8905  & 0.9272 & 0.0447 & 0.8901 \\
%                 & level 4 & 0.8967 & 0.0631 & 0.8844  & 0.9161 & 0.0578 & 0.8850  \\
%                 & level 5 & 0.8759 & 0.0870 & 0.8649 & 0.8969 & 0.0797 &  0.8686 \\
%                \hline
		\end{tabular}}
	\end{center}
	\label{num3}
\end{table*}

\section{Conclusion}

We propose an RGB-D SOD network equipped with cross-modality feature modulation and adaptive feature selection.
The former effectively integrates the multi-modality complementarities while the latter adaptively highlights saliency-related features.
%
%Coupled with a bottom-up inference, our network can yield accurate and edge-preserving saliency detection results.
%
We demonstrate that both elaborate integration of cross-modality features  and adaptive selection of multi-modality spatial and channel features can boost the performance of SOD. Experiment results also demonstrate that our method achieves new state-of-the-art performance on six benchmarks.

\section*{Acknowledgments}
This research was supported by SenseTime-NTU Collaboration Project, Singapore MOE AcRF Tier 1 (2018-T1-002-056), NTU NAP, in part by the Fundamental Research Funds for the Central Universities under Grant 2019RC039, and in part by China Postdoctoral Science Foundation Grant 2019M660438.

%\clearpage
\bibliographystyle{splncs04}
\bibliography{eccv}

\begin{thebibliography}{10}
\providecommand{\url}[1]{\texttt{#1}}
\providecommand{\urlprefix}{URL }
\providecommand{\doi}[1]{https://doi.org/#1}

\bibitem{cross-entropy}
Boer, P.T.D., Kroese, D.P., Mannor, S., Rubinstein, R.Y.: A tutorial on the
  cross-entropy method. Annals of Operations Research  \textbf{134}(1),  19--67
  (2005)

\bibitem{Fmeasure}
Borji, A., Cheng, M.M., Jiang, H., Li, J.: Salient object detection: A
  benchmark. IEEE Trans. Image Process.  \textbf{24}(12),  5706--5722 (2015)

\bibitem{PCFN}
Chen, H., Li, Y.: Progressively complementarity-aware fusion network for
  {RGB-D} salient object detection. In: CVPR. pp. 3051--3060 (2018)

\bibitem{TAN}
Chen, H., Li, Y.: Three-stream attention-aware network for {RGB-D} salient
  object detection. IEEE Trans. Image Process.  \textbf{28}(6),  2825--2835
  (2019)

\bibitem{DCFF}
Chen, H., Li, Y., Su, D.: Discriminative cross-modal transfer learning and
  densely cross-level feedback fusion for {RGB-D} salient object detection.
  IEEE Trans. Cybern. pp. 1--13 (2019)

\bibitem{MMCI}
Chen, H., Li, Y., Su, D.: Multi-modal fusion network with multiscale multi-path
  and cross-modal interactions for {RGB-D} salient object detection. Pattern
  Recognit.  \textbf{86},  376--385 (2019)

\bibitem{Spattention}
Chen, L., Zhang, H., Xiao, J., Nie, L., Shao, J., Liu, W., Chua, T.S.:
  {SCA-CNN}: Spatial and channel-wise attention in convolutional networks for
  image captioning. In: CVPR. pp. 5659--5667 (2017)

\bibitem{RERVIEW}
Cong, R., Lei, J., Fu, H., Cheng, M.M., Lin, W., Huang, Q.: Review of visual
  saliency detectioin with comprehensive information. IEEE Trans. Circuits
  Syst. Video Technol  \textbf{29}(10),  2941--2959 (2019)

\bibitem{crm2019tc}
Cong, R., Lei, J., Fu, H., Hou, J., Huang, Q., Kwong, S.: Going from {RGB} to
  {RGBD} saliency: A depth-guided transformation model. IEEE Trans. Cybern. pp.
  1--13 (2019)

\bibitem{DCMC}
Cong, R., Lei, J., Zhang, C., Huang, Q., Cao, X., Hou, C.: Saliency detection
  for stereoscopic images based on depth confidence analysis and multiple cues
  fusion. IEEE Signal Process. Lett.  \textbf{23}(6),  819--823 (2016)

\bibitem{S-measure}
Fan, D.P., Cheng, M.M., Liu, Y., Li, T., Borji, A.: Structure-measure: A new
  way to evaluate foreground maps. In: ICCV. pp. 4548--4557 (2017)

\bibitem{fan2020bbs}
Fan, D.P., Zhai, Y., Borji, A., Yang, J., Shao, L.: {BBS-Net}: {RGB-D} salient
  object detection with a bifurcated backbone strategy network. In: ECCV (2020)

\bibitem{LBE}
Feng, D., Barnes, N., You, S., McCarthy, C.: Local background enclosure for
  {RGB-D} salient object detection. In: CVPR. pp. 2343--2350 (2016)

\bibitem{AFNet}
Feng, M., Lu, H., Ding, E.: Attentive feedback network for boundary-aware
  salient object detection. In: CVPR. pp. 1623--1632 (2019)

\bibitem{DualAttention}
Fu, J., Liu, J., Tian, H., Li, Y.: Dual attention network for scene
  segmentation. In: CVPR. pp. 3146--3154 (2019)

\bibitem{Fu2020JLDCF}
Fu, K.F., Fan, D.P., Ji, G.P., Zhao, Q.: {JL-DCF}: Joint learning and
  densely-cooperative fusion framework for {RGB-D} salient object detection.
  In: CVPR. pp. 3052--3062 (2020)

\bibitem{Guan2018}
Guan, W., Wang, T., Qi, J., Zhang, L., Lu, H.: Edge-aware convolutional neural
  network based salient object detection. IEEE Signal Process. Lett. pp. 114 --
  118 (2018)

\bibitem{CTMF}
Han, J., Chen, H., Liu, N., Yan, C., Li, X.: {CNN}s-based {RGB-D} saliency
  detection via cross-view transfer and multiview fusion. IEEE Trans. Cybern.
  \textbf{48}(11),  3171--3183 (2018)

\bibitem{DSS}
Hou, Q., Cheng, M.M., Hu, X., Borji, A., Tu, Z., Torr, P.H.: Deeply supervised
  salient object detection with short connections. IEEE Trans. Pattern Anal.
  Mach. Intell.  \textbf{41}(4),  815--828 (2019)

\bibitem{ChannelAtt}
Hu, J., Shen, L., Sun, G.: Squeeze-and-excitation networks. In: CVPR. pp.
  7132--7141 (2018)

\bibitem{ACSD}
Ju, R., Liu, Y., Ren, T., Ge, L., Wu, G.: Depth-aware salient object detection
  using anisotropic center-surround difference. Signal Process.: Image Commun.
  \textbf{38},  115--126 (2015)

\bibitem{ASIF-Net}
Li, C., Cong, R., Kwong, S., Hou, J., Fu, H., Zhu, G., Zhang, D., Huang, Q.:
  {ASIF-Net}: Attention steered interweave fusion network for {RGBD} salient
  object detection. IEEE Trans. Cybern. pp. 1--13 (2020)

\bibitem{SSD}
Li, G., Zhu, C.: A three-pathway psychobiological framework of salient object
  detection using stereoscopic technology. In: ICCVW. pp. 3008--3014 (2017)

\bibitem{LFSD}
Li, N., Ye, J., Ji, Y., Ling, H., Yu, J.: Saliency detection on light field.
  In: CVPR. pp. 2806--2813 (2014)

\bibitem{DSR}
Li, X., Lu, H., Zhang, L., Ruan, X., Yang, M.H.: Saliency detection via dense
  and sparse reconstruction. In: ICCV. pp. 2976--2983 (2013)

\bibitem{PoolNet}
Liu, J., Hou, Q., Cheng, M.M., Feng, J., Jiang, J.: A simple pooling-based
  design for real-time salient object detection. In: CVPR. pp. 3917--3926
  (2019)

\bibitem{Niu2012}
Niu, Y., Geng, Y., Li, X., Liu, F.: Leveraging stereopsis for saliency
  analysis. In: CVPR. pp. 454--461 (2012)

\bibitem{TADAM}
Oreshkin, B.N., Rodriguez, P., Lacoste, A.: {TADAM}: Task dependent adaptive
  metric for improved few-shot learning. In: NeurIPS. pp. 721--731 (2018)

\bibitem{SMD}
Peng, H., Li, B., Ling, H., Hu, W., Xiong, W., Maybank, S.J.: Salient object
  detection via structured matrix decomposition. IEEE Trans. Pattern Anal.
  Mach. Intell.  \textbf{39}(4),  818--832 (2017)

\bibitem{Peng2014}
Peng, H., Li, B., Xiong, W., Hu, W., Ji, R.: {RGBD} salient object detection: A
  benchmark and algorithms. In: ECCV. pp. 92--109 (2014)

\bibitem{FiLM}
Perez, E., Strub, F., de~Vries, H., Dumoulin, V., Courville, A.: {FiLM}: Visual
  reasoning with a general conditioning layer. In: AAAI. pp. 3942--3951 (2018)

\bibitem{DMRA}
Piao, Y., Ji, W., Li, J., Zhang, M., Lu, H.: Depth-induced multi-scale
  recurrent attention network for saliency detection. In: ICCV. pp. 7254--7263
  (2019)

\bibitem{A2dele}
Piao, Y., Rong, Z., Zhang, M., Ren, W., Lu, H.: A2dele: Adaptive and attentive
  depth distiller for efficient {RGB-D} salient object detection. In: CVPR. pp.
  9060--9069 (2020)

\bibitem{BASNet}
Qin, X., Zhang, Z., Huang, C., Gao, C., Dehghan, M., Jagersand, M.: {BASNet}:
  Boundary-aware salient object detection. In: CVPR. pp. 7479--7489 (2019)

\bibitem{DF}
Qu, L., He, S., Zhang, J., Tian, J., Tang, Y., Yang, Q.: {RGBD} salient object
  detection via deep fusion. IEEE Trans. Image Process.  \textbf{26}(5),
  2274--2285 (2017)

\bibitem{VGG}
Simonyan, K., Zisserman, A.: Very deep convolutional networks for large-scale
  image recognition. arXiv preprint arXiv:1409.1556  (2014)

\bibitem{Song2017}
Song, H., Liu, Z., Du, H., Sun, G., Le~Meur, O., Ren, T.: Depth-aware salient
  object detection and segmentation via multiscale discriminative saliency
  fusion and bootstrap learning. IEEE Trans. Image Process.  \textbf{26}(9),
  4204--4216 (2017)

\bibitem{Attention17}
Vaswani, A., Shazeer, N., Parmar, N., Uszkoreit, J., Jones, L., Gomez, A.N.,
  Kaiser, L.: Attention is all you need. In: NeurIPS. pp. 5998--6008 (2017)

\bibitem{SR2018}
Wang, X., Yu, K., Dong, C., Loy, C.C.: Recovering realistic texture in image
  super-resolution by deep spatial feature transform. In: CVPR. pp. 606--615
  (2018)

\bibitem{MultiAttention}
Yu, D., Fu, J., Mei, T., Rui, Y.: Multi-level attention networks for visual
  question answering. In: CVPR. pp. 4709--4717 (2017)

\bibitem{RCRR}
Yuan, Y., Li, C., Kim, J., Cai, W., Feng, D.D.: Reversion correction and
  regularized random walk ranking for saliency detection. IEEE Trans. Image
  Process.  \textbf{27}(3),  1311--1322 (2018)

\bibitem{Zhang2020CVPR}
Zhang, J., Fan, D.P., Dai, Y., Anwar, S., Saleh, F.S., Zhang, T., Barnes, N.:
  {UC-Net}: Uncertainty inspired {RGB-D} saliency detection via conditional
  variational autoencoders. In: CVPR. pp. 8582--8591 (2020)

\bibitem{Zhangmiao2020CVPR}
Zhang, M., Ren, W., Piao, Y., Rong, Z., Lu, H.: Select,supplement and focus for
  rgb-d saliency detection. In: CVPR. pp. 3472--3481 (2020)

\bibitem{ChannelAttention}
Zhang, Y., Li, K., Li, K., Wang, L., Zhong, B., Fu, Y.: Image super-resolution
  using very deep residual channel attention networks. In: ECCV. pp. 286--301
  (2018)

\bibitem{CPFP}
Zhao, J., Cao, Y., Fan, D.P., Cheng, M.M., Li, X.Y., Zhang, L.: Contrast prior
  and fluid pyramid integration for {RGBD} salient object detection. In: CVPR.
  pp. 3927--3936 (2019)

\bibitem{EGNet}
Zhao, J., Liu, J., Fan, D.P., Cao, Y., Yang, J., Cheng, M.M.: {EGNet}: Edge
  guidance network for salient object detection. In: ICCV. pp. 8779--8788
  (2019)

\bibitem{Zhu2018}
Zhu, C., Li, G.: A multilayer backpropagation saliency detection algorithm and
  its applications. Multimedia Tools Appl.  \textbf{77}(19),  25181--25197
  (2018)

\end{thebibliography}
\end{document}